\documentclass[10pt,twocolumn,letterpaper]{article}

\usepackage{cvpr}     

\usepackage{times}
\usepackage{graphicx}
\usepackage{amsmath}
\usepackage{amssymb}
\usepackage{booktabs}
\usepackage{multirow}
\usepackage[table]{xcolor}    
\usepackage{wrapfig}
\usepackage{enumitem}
\usepackage{dblfloatfix}       
\usepackage[colorlinks,linkcolor=blue,citecolor=blue,urlcolor=blue,breaklinks]{hyperref}  

\begin{document}

\title{Compatibility of Face Embeddings Across Deep Neural Networks}

\author{Fizza Rubab\\
Michigan State University\\
{\tt\small rubabfiz@msu.edu}
\and
Yiying Tong\\
Michigan State University\\
{\tt\small ytong@msu.edu}
\and
Arun Ross\\
Michigan State University\\
{\tt\small rossarun@msu.edu}
}

\twocolumn[{%
\renewcommand\twocolumn[1][]{#1}%
{\centering\small Paper accepted at IEEE/IAPR International Joint Conference on Biometrics (IJCB), September 2026.\par}
\vspace{-0.6cm}
\maketitle
\includegraphics[width=\textwidth]{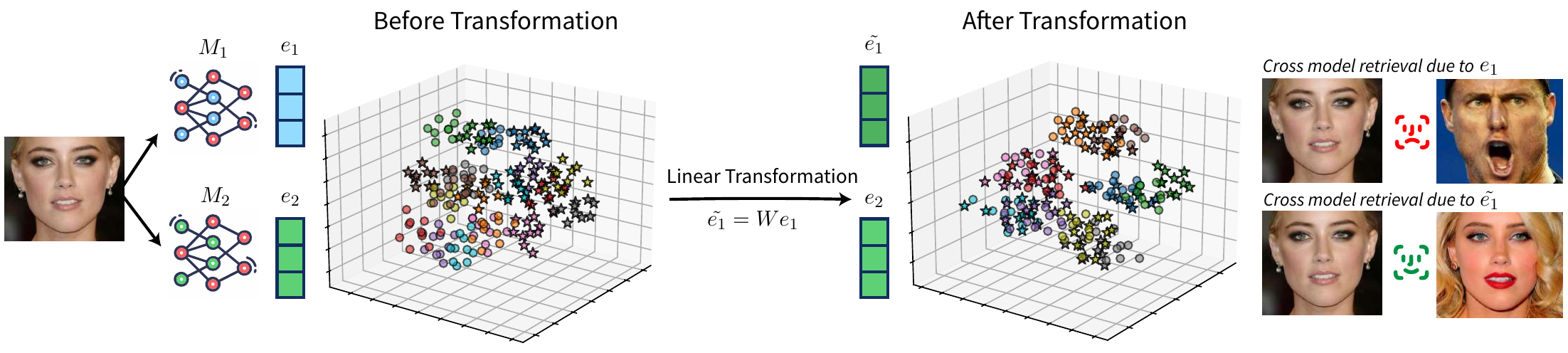}
\captionof{figure}{\textbf{Linear alignment of embedding spaces.} Two independently trained models ($M_1$ and $M_2$) produce distinct embeddings ($e_1$ and $e_2$) for the same face image. A simple linear transformation, $W$, aligns embeddings in identity space, thereby improving cross-model identification accuracy.}
\label{fig:overview}
}]


    \begin{abstract} Automated face recognition has made rapid strides over the past decade due to the unprecedented rise of deep neural network (DNN) models that can be trained for domain-specific tasks. At the same time, large foundation models that are pretrained on broad vision or vision-language tasks have shown impressive generalization across diverse domains, including biometrics. This raises an important question: \textit{Do different DNN models---both domain-specific and foundation models---encode facial identity in similar ways, despite being trained on different datasets, loss functions, and architectures?} In this regard, we directly analyze the geometric structure of embedding spaces imputed by different DNN models. Treating embeddings of face images as point clouds, we study whether simple affine transformations can align face representations of one model with another. Our findings reveal substantial cross-model compatibility: low-capacity \emph{linear} mappings substantially improve cross-model face recognition over unaligned baselines for both identification and verification, including across foundation models never trained for face recognition. Alignment patterns generalize across datasets and vary systematically across model families, indicating representational convergence in facial identity encoding. These findings reframe independently trained templates as transferable rather than revocable, with implications for interoperability, ensemble design, and biometric template security.

\end{abstract}
\section{Introduction}
Modern face recognition systems rely on deep neural network (DNN) models that map face images to compact embedding vectors. The similarity or dissimilarity between two face images is then measured through distances or inner products between their corresponding embedding vectors. Domain-specific models such as ArcFace~\cite{deng2019arcface}, AdaFace~\cite{kim2022adaface}, and MagFace~\cite{meng2021magface} have been trained only on face images and achieve high accuracy. More recently, foundation models, which have been pre-trained on a broad range of data and tasks, have been evaluated in the context of face recognition~\cite{sony2025benchmarking, sony2025foundation}. These models (both domain-specific and foundation) produce distinct embedding spaces learned under different architectures, data regimes, and training objectives.

Prior work suggests that these embedding spaces may be fundamentally incompatible. Bhatta et al.~\cite{bhatta2025deep, bhatta2026revocable} show that independently trained instances of the \textit{same} DNN architecture yield mutually incompatible embeddings, where the embeddings of the same face image corresponding to two different models are radically different, an effect interpreted as implicit template revocability. However, this warrants closer examination. While models are not trained on identical images or even identities, they share structural learning objectives: face-specific DNNs separate identities through metric learning, while foundation models encode visual structure through self-supervision or vision-language alignment. Both encourage structured organization of the identity manifold. We therefore investigate whether observed incompatibility reflects fundamentally different representations or simply misaligned parameterizations of a common underlying structure.

\begin{figure*}[t]
    \centering
  \includegraphics[width=0.75\textwidth]{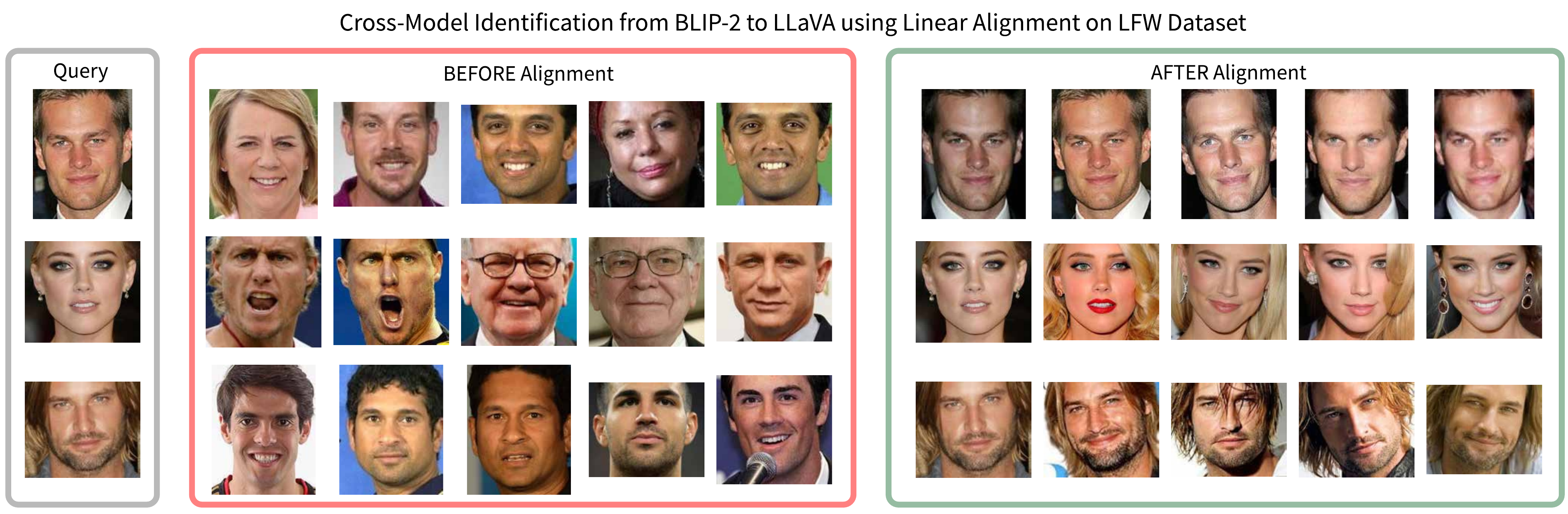}
  \caption{\textbf{Qualitative cross-model face identification results.} Query images with BLIP-2 embeddings (left) yield incorrect nearest neighbors with LLaVA embeddings before alignment (center), but correct identity matches after linear alignment (right).}
  \label{fig:gallery}
  \vspace{-6pt}
\end{figure*}

This motivates our research question: \textit{Do different DNNs encode facial identity differently, or are there commonalities between them?} Given embeddings from two models on the same set of identities/images, we learn simple affine mappings between embedding spaces and evaluate cross-model face identification as demonstrated in Figure~\ref{fig:overview}. We study three low-capacity transformations: Procrustes alignment, unconstrained linear regression, and Ridge regression, and measure performance on both identification and verification tasks across face-specific and foundation models.

Our results reveal substantial cross-model compatibility. Linear mappings substantially improve identification accuracy over unaligned baselines, and transformations generalize across datasets. This suggests representational convergence: independently trained models often differ more in coordinate parameterization than intrinsic identity geometry, with implications for interoperability, ensemble design, and biometric template security.\footnote{Code: \url{https://github.com/Fizza-Rubab/face-embedding-compatibility}}

\noindent\textbf{Contributions.} Our main contributions are:
\begin{enumerate}[noitemsep,topsep=0pt]
    \item A systematic study of face embedding compatibility across both domain-specific face recognition models and general-purpose foundation models.
    \item Evidence that low-capacity linear transformations enable substantial cross-model identification accuracy, indicating shared geometric identity structure.
    \item Demonstration of cross-dataset generalization, showing that learned alignment transformations transfer across datasets.
    \item A structural analysis of cross-model compatibility, revealing hierarchical model organization and directional asymmetry in target spaces.
    \item A geometric interpretation grounded in manifold learning theory, explaining why simple linear transformations suffice via local tangent-space alignment.
\end{enumerate}

\section{Related Work}

\textbf{Face Recognition with DNNs.} Face-specific models trained with metric learning objectives (e.g., FaceNet~\cite{schroff2015facenet}, ArcFace~\cite{deng2019arcface}, CosFace~\cite{wang2018cosface}, AdaFace~\cite{kim2022adaface}, MagFace~\cite{meng2021magface}) achieve state-of-the-art accuracy by optimizing angular margin losses. Foundation models pretrained on generic vision or vision-language tasks have also been explored for face recognition~\cite{sony2025benchmarking,chettaoui2024face}, demonstrating promising zero-shot performance and competitive accuracy with fine-tuning. Our work differs by studying \textit{cross-model} alignment: whether embeddings from independently trained models can be aligned through simple transformations.\\

\noindent\textbf{Representation Similarity and Alignment.} 
Understanding and comparing learned representations in DNNs has been extensively  studied. Klabunde et al.~\cite{klabunde2025similarity} provide a taxonomy of functional and representational similarity analysis (RSA) methods. Techniques such as CKA~\cite{kornblith2019similarity}, SVCCA~\cite{raghu2017svcca}, and convergent learning analyses~\cite{li2015convergent} quantify similarity through internal activations in DNNs. In cognitive neuroscience research, recent work using RSA with naturalistic dynamic face stimuli showed that face-trained DNNs exhibit consistent representational geometry across architectures and partially align with human behavioral judgments and neural responses~\cite{jiahui2023modeling}. RSA has also inspired cross-lingual word embedding alignment, where embeddings of independently trained DNNs share geometric structure alignable through linear transformations. Conneau et al.~\cite{conneau2017word} used orthogonal Procrustes for aligning word embedding spaces, while Artetxe et al.~\cite{artetxe2018robust} developed unsupervised methods. We extend this alignment perspective to vision by operating directly on final face embeddings across independently trained models, restricting ourselves to simple linear transformations for their simplicity and geometric interpretability.\\

\noindent\textbf{Biometric Template Security.} Biometric templates (i.e., embeddings) are typically irreplaceable once compromised~\cite{jain2008biometric}, motivating the development of template protection schemes~\cite{ratha2007generating}. However, protected templates remain vulnerable to reconstruction attacks~\cite{otroshi2024}. Furthermore, Bhatta et al.~\cite{bhatta2025deep} demonstrated that training multiple instances of the same architecture yields mutually incompatible embeddings, arguing this provides inherent template revocability. Our findings complicate this picture: face embeddings from different DNNs can be registered with high accuracy through linear transformations. An adversary with probe images from two systems and embeddings could estimate alignment transformations and match templates across applications, suggesting revocability is more fragile than previously assumed.

\section{Models}
\label{sec:models}

We extract face embeddings using two categories of deep neural networks: \textit{face-specific models} trained specifically on face images for identity discrimination and general-purpose \textit{foundation models} pretrained on large-scale vision or vision-language data. 

Face-specific models are trained on large-scale labeled face datasets using learning objectives that enforce intra-class compactness and inter-class separation. Over the years, a large number of face recognition techniques have been developed~\cite{kim202650}. ArcFace~\cite{deng2019arcface} introduces additive angular margins, AdaFace~\cite{kim2022adaface} adapts margins based on image quality, MagFace~\cite{meng2021magface} incorporates magnitude-aware regularization, and KPRPE~\cite{kim2024keypoint} leverages facial keypoints to improve invariance to scale, translation, and pose.  

Foundation models are trained on large-scale image, image–text, or mask-annotated datasets with objectives that align visual and textual features or learn general-purpose semantic representations rather than directly optimizing for identity discrimination. In this work, we evaluate models spanning contrastive learning (CLIP~\cite{radford2021clip}, ALIGN~\cite{jia2021align}), self-supervised representation learning (DINOv2~\cite{oquab2023dinov2}), segmentation-based pretraining (SAM~\cite{kirillov2023sam}), dense prediction (Florence-2~\cite{xiao2024florence}) and multimodal grounding or instruction tuning (BLIP-2~\cite{li2023blip2}, LLaVA~\cite{liu2023llava}, Kosmos-2~\cite{peng2023kosmos}, InternVL3~\cite{zhu2025internvl3}). Table~\ref{tab:models} summarizes the models used in this work and their properties.

\begin{table}[h]
\centering
\caption{Summary of the DNN models whose embeddings were analyzed in this study.}
\label{tab:models}
\resizebox{\columnwidth}{!}{%
\begin{tabular}{llllllr}
\toprule
\textbf{Category} & \textbf{Model} & \textbf{Variant} & \textbf{Architecture} & \textbf{Training Data} & \textbf{Training Objective} & \textbf{Dimension} \\
\midrule
\multirow{4}{*}{\parbox{2cm}{Face-Specific}} 
& ArcFace   & ir101              & iResNet-101 & WebFace4M & Additive angular margin    & 512  \\
& AdaFace   & ir101              & iResNet-101 & MS1MV2 & Adaptive margin            & 512  \\
& MagFace   & ir100             & iResNet-100 & MS1MV2 & Magnitude-aware margin    & 512  \\
& KPRPE     & vit-base              & ViT-Base & WebFace4M & Keypoint-guided relational & 512  \\
\midrule
\multirow{10}{*}{\parbox{2cm}{Foundation \\ Models}} 
& CLIP      & ViT-B/32          & ViT        & 400M image-text & Vision-language contrastive  & 512  \\
& ALIGN     & align-base        & EfficientNet & 1B image-text & Vision-language contrastive & 640  \\
& DINOv2    & ViT-B/14          & ViT        & LVD-142M & Self-supervised distillation    & 768  \\
& SAM       & ViT-B             & ViT        & SA-1B & Supervised segmentation    & 256  \\
& BLIP-2    & blip2-opt-2.7b    & ViT + LLM  & Web data & Vision-language generative   & 1,408 \\
& LLaVA     & llava-1.5-7b      & ViT + LLM  & Multimodal & Vision-language instruction  & 1,024 \\
& Kosmos-2  & kosmos-2-patch14-224 & ViT + LLM & Web data & Grounded vision-language & 1,024 \\
& InternVL3 & InternVL3-1B      & ViT + LLM  & Diverse & Vision-language contrastive  & 1,024  \\
& Florence-2 & Florence-2-base  & ViT + LLM &  FLD-5B & Seq2Seq vision-language modeling & 768  \\
& ViT       & vit-base-patch16-224 & ViT     & ImageNet-21K & Supervised classification  & 768  \\
\bottomrule
\end{tabular}%
}
\end{table}


\section{Datasets}
We evaluate our method on three widely used face recognition benchmarks with varying characteristics. Celebrities in Frontal-Profile (CFP)~\cite{sengupta2016cfp} contains 500 identities with frontal-profile image pairs designed to evaluate pose-invariant recognition. We use only the 10 frontal images per identity, yielding 5,000 images total in a controlled setting with consistent acquisition conditions. Labeled Faces in the Wild (LFW)~\cite{huang2008labeled} comprises 5,749 identities with 13,233 unconstrained images collected from news articles, representing in-the-wild conditions with natural pose and lighting variation. We use center-cropped images at $160\!\!\times\!\!160$ resolution to remove uninformative border regions. CASIA-WebFace~\cite{yi2014learning} provides 10,575 identities with 494,414 images scraped from the web, offering large-scale diversity in ethnicity, age, and image quality. To maintain computational tractability for all our experiments, we randomly sample 5 images per identity from CASIA-WebFace, yielding 52,875 images for our experiments.

\section{Methodology}

\subsection{Problem Formulation}

Let $\smash{\mathcal{F} = \{f_1, f_2, \ldots, f_M\}}$ be a set of $M$ DNN models, where each $\smash{f_m : \mathcal{I} \rightarrow \mathbb{R}^{d_m}}$ maps a face image (in the set of possible images $\mathcal{I}$) to a point in $\mathbb{R}^{d_m}$ through a $d_m$-dimensional embedding. Given a shared set of $N$ face images $\{I_i\}_{i=1}^{N}$ with identity labels $\{y_i\}_{i=1}^{N}$, model $f_m$ produces an embedding matrix $\smash{\mathbf{X}^{\scriptscriptstyle(m)} \in \mathbb{R}^{N \times d_m}}$, where its $i$-th row, $\mathbf{x}^{\scriptscriptstyle(m)}_i$, is the embedding of image $I_i$.

For a pair of models $(f_a, f_b)$, we ask: does there exist a transformation $\smash{\varphi : \mathbb{R}^{d_a} \!\rightarrow\! \mathbb{R}^{d_b}}$ such that $\smash{\varphi(\mathbf{x}_i^{\scriptscriptstyle(a)}) \approx \mathbf{x}_i^{\scriptscriptstyle(b)}}$? We study this through the lens of \textit{alignment}: estimating $\varphi$ from a subset of identity-disjoint training pairs and evaluating whether identity relationships transfer to the remaining held-out identities.

\subsection{Preprocessing}

After extraction, embeddings from each model are $\ell_2$-normalized to unit length. We center each embedding matrix using training set statistics and zero-pad them to a common dimension $\smash{D \!=\! \max(d_a, d_b)}$. Formally, given training embeddings $\smash{\mathbf{X}_{\text{tr}} \in\! \mathbb{R}^{n \times d_a}}$ and $\smash{\mathbf{Y}_{\text{tr}} \in\! \mathbb{R}^{n \times d_b}}$, we compute their respective means $\smash{\boldsymbol{\mu}_X \!=\! \frac{1}{n}\sum_i \mathbf{x}_i}$ and $\smash{\boldsymbol{\mu}_Y \!=\! \frac{1}{n}\sum_i \mathbf{y}_i}$, then center and pad:

\begin{align} \tilde{\mathbf{X}} = \left[\mathbf{X} - \boldsymbol{\mu}_X \;\Big|\; \mathbf{0}_{D-d_a}\right] \in \mathbb{R}^{N \times D}, \quad \\\tilde{\mathbf{Y}} = \left[\mathbf{Y} - \boldsymbol{\mu}_Y \;\Big|\; \mathbf{0}_{D-d_b}\right] \in \mathbb{R}^{N \times D}. \end{align}

Test embeddings are centered using training statistics (with $n<N$) to prevent data leakage by ensuring the transformation does not incorporate any information about test identities.

\subsection{Alignment Methods}

We study three alignment methods of increasing expressiveness, each learning a linear map $\mathbf{W} \in \mathbb{R}^{D \times D}$ from source to target space.

\vspace{0.5em} \noindent\textbf{Procrustes Alignment.} The orthogonal Procrustes problem seeks a rotation matrix $\mathbf{W}^*$ that maps training samples $\tilde{\mathbf{X}}_{\text{tr}}$ to $\tilde{\mathbf{Y}}_{\text{tr}}$:

\begin{equation} \mathbf{W}^* = \arg\!\min_{\mathbf{W}} \left\| \tilde{\mathbf{X}}_{\text{tr}} \mathbf{W} - \tilde{\mathbf{Y}}_{\text{tr}} \right\|_F^2, \;\mathrm{s.t.}\; \mathbf{W}^\top \mathbf{W}=\mathbf{I}, \end{equation}
where, $\mathbf{I}$ is the $D$-dimensional identity matrix.

This admits a closed-form solution~\cite{schonemann1966generalized} via singular value decomposition (SVD). Let $\mathbf{M} = \tilde{\mathbf{X}}_{\text{tr}}^\top \tilde{\mathbf{Y}}_{\text{tr}}$. Given the SVD of $\mathbf{M} = \mathbf{U}\boldsymbol{\Sigma}\mathbf{V}^\top$ as the product of two orthogonal matrices $\mathbf{U}$, $\mathbf{V}^\top \in O(D)$, and a diagonal matrix $\smash{\boldsymbol{\Sigma}=\mathrm{diag}[\sigma_1,\dots,\sigma_D]}$, the optimal solution is given by
\begin{equation} \mathbf{W}^* = \mathbf{U}\mathbf{V}^\top. \end{equation}

The aligned embedding is $\hat{\mathbf{X}} = \tilde{\mathbf{X}} \mathbf{W}^*$. The orthogonality constraint preserves pairwise angles and relative distances, making this the most metric-preserving alignment.

\vspace{0.5em} \noindent\textbf{Linear Alignment.} When we relax the orthogonality constraint, and seek an unconstrained linear map instead:

\begin{equation} \mathbf{W}^* = \arg\min_{\mathbf{W}} \left\| \tilde{\mathbf{X}}_{\text{tr}} \mathbf{W} - \tilde{\mathbf{Y}}_{\text{tr}} \right\|_F^2. \end{equation}

This also has a closed-form solution in the least-squares sense, through the pseudo-inverse of $\tilde{\mathbf{X}}_{\text{tr}}$,
\begin{equation} \mathbf{W}^* = \left(\tilde{\mathbf{X}}_{\text{tr}}^\top \tilde{\mathbf{X}}_{\text{tr}}\right)^{-1} \tilde{\mathbf{X}}_{\text{tr}}^\top \tilde{\mathbf{Y}}_{\text{tr}}, \end{equation}
which allows for anisotropic scaling and shearing in addition to rotation.

\vspace{0.5em} \noindent\textbf{Ridge Alignment.} To prevent overfitting when training samples are limited, we can include an $\ell_2$ regularization with a weight $\alpha>0$:
\begin{equation} \mathbf{W}^* = \arg\min_{\mathbf{W}} \left\| \tilde{\mathbf{X}}_{\text{tr}} \mathbf{W} - \tilde{\mathbf{Y}}_{\text{tr}} \right\|_F^2 + \alpha \|\mathbf{W}\|_F^2 ,\end{equation}
resulting in a damped least-squares solution $\smash{\mathbf{W}^* = \left(\tilde{\mathbf{X}}_{\text{tr}}^\top \tilde{\mathbf{X}}_{\text{tr}} + \alpha \mathbf{I}\right)^{-1} \tilde{\mathbf{X}}_{\text{tr}}^\top \tilde{\mathbf{Y}}_{\text{tr}}}$.

\section{Experimental Setup}
We evaluate cross-model embedding alignment across two complementary dimensions: \textit{task type} and \textit{generalization setting}. For tasks, we assess performance on face identification and face verification. For generalization, we consider both intra-dataset alignment and cross-dataset alignment. Finally, we analyze alignment across two \textit{model categories}: face-specific models and generic foundation models, as described in Section~\ref{sec:models}.  For brevity, we show representative experiments and results in the main text; complete tables and configurations for all model pairs appear in the supplementary material. In the descriptions that follow, the notation $model\; A \rightarrow model\; B$ indicates that embeddings from $A$ are aligned to embeddings from $B$.\\

{\textbf{Face Identification.}} Identification compares a query embedding with all gallery embeddings (in the database) and retrieves those with highest similarity. For a query $\mathbf{x}_i$ and gallery $\{\mathbf{y}_j\}_{j=1}^{N}$, we compute $\ell_2$-normalized cosine similarities and rank gallery items by decreasing similarity. We report Rank-1 accuracy and mean Average Precision (mAP), and plot Cumulative Matching Characteristic (CMC) curves showing recognition rate across ranks ($k = 1, \ldots, 50$). For intra-dataset identification, we use 70\% of identities for training the alignment and 30\% for evaluation, ensuring no identity appears in both sets. Results are averaged over five random seeds with different partitions. For cross-dataset identification, alignment is derived using all identities from the source dataset and evaluated on all identities in the target dataset.\\

{\textbf{Face Verification.}} Verification evaluates whether two images depict the same identity or not. Given image pairs labeled as genuine (same identity) or impostor (different identities), we threshold similarity scores to make accept/reject decisions. We report Area Under the ROC Curve (AUC) and True Match Rate at False Match Rate of 1\% (TMR@1\%FMR) in the main tables; Equal Error Rate (EER) and complete per-pair results appear in the supplementary material. For intra-dataset verification, alignment is derived using 70\% of identities (same split as identification). We then generate all possible genuine pairs from the remaining 30\% test identities and sample an equal number of impostor pairs uniformly at random, maintaining a balanced 1:1 ratio. For cross-dataset verification, alignment is derived using all identities from the source dataset. For evaluation, we sample 10,000 genuine and 10,000 impostor pairs from all identities in the target dataset for computational efficiency. Both protocols average results over five random seeds.

\section{Results}
\begin{table*}[t]
\centering
\caption{\textbf{Intra-dataset cross-model results.} Identification (Rank-1, mAP) and verification (AUC, TMR@1\%FMR). Top block: face-specific model pairs; bottom block: foundation-model pairs. \emph{Baseline} is the unaligned cross-model comparison.}
\label{tab:intra_main}
\resizebox{\textwidth}{!}{%
\begin{tabular}{ll *{12}{c}}
\toprule
\textbf{Model Pair} & \textbf{Method} & \multicolumn{4}{c}{\textbf{CFP}} & \multicolumn{4}{c}{\textbf{LFW}} & \multicolumn{4}{c}{\textbf{WebFace}} \\
\cmidrule(lr){3-6}\cmidrule(lr){7-10}\cmidrule(lr){11-14}
 & & R-1 & mAP & AUC & TMR & R-1 & mAP & AUC & TMR & R-1 & mAP & AUC & TMR \\
\midrule
\rowcolor{gray!15} \multirow{4}{*}{ArcFace $\rightarrow$ AdaFace} & Baseline & 0.57 ± 0.09 & 0.09 ± 0.01 & 0.50 ± 0.02 & 1.20 ± 0.18 & 0.20 ± 0.17 & 0.09 ± 0.01 & 0.55 ± 0.06 & 0.80 ± 0.18 & 0.03 ± 0.01 & 0.08 ± 0.01 & 0.49 ± 0.00 & 0.87 ± 0.05 \\
 & Procrustes & \textbf{98.53 ± 0.38} & \textbf{0.80 ± 0.01} & 0.96 ± 0.00 & \textbf{68.41 ± 2.31} & \textbf{99.62 ± 0.10} & 0.99 ± 0.00 & \textbf{0.99 ± 0.00} & 96.73 ± 0.48 & \textbf{99.96 ± 0.01} & 0.98 ± 0.00 & \textbf{0.95 ± 0.00} & 89.41 ± 0.17 \\
 & Linear & 97.64 ± 0.26 & 0.77 ± 0.01 & 0.96 ± 0.00 & 62.34 ± 3.48 & 99.62 ± 0.09 & \textbf{0.99 ± 0.00} & 0.99 ± 0.00 & \textbf{96.84 ± 0.45} & 99.96 ± 0.01 & \textbf{0.98 ± 0.00} & 0.95 ± 0.00 & \textbf{89.47 ± 0.19} \\
 & Ridge & 97.68 ± 0.50 & 0.77 ± 0.01 & \textbf{0.96 ± 0.00} & 63.00 ± 3.27 & 99.61 ± 0.10 & 0.99 ± 0.00 & 0.99 ± 0.00 & 96.83 ± 0.46 & 99.96 ± 0.01 & 0.98 ± 0.00 & 0.95 ± 0.00 & 89.46 ± 0.19 \\
\cmidrule{1-14}
\rowcolor{gray!15} \multirow{4}{*}{AdaFace $\rightarrow$ KPRPE} & Baseline & 0.77 ± 0.18 & 0.10 ± 0.01 & 0.50 ± 0.01 & 1.21 ± 0.42 & 0.25 ± 0.27 & 0.10 ± 0.04 & 0.49 ± 0.10 & 0.98 ± 0.71 & 0.02 ± 0.00 & 0.08 ± 0.00 & 0.50 ± 0.00 & 0.82 ± 0.05 \\
 & Procrustes & 77.99 ± 1.27 & 0.69 ± 0.01 & 0.97 ± 0.00 & 69.32 ± 2.54 & 98.84 ± 0.19 & 0.99 ± 0.00 & \textbf{0.99 ± 0.00} & 97.09 ± 0.46 & 99.87 ± 0.01 & 0.97 ± 0.00 & 0.95 ± 0.00 & 88.80 ± 0.21 \\
 & Linear & \textbf{83.72 ± 0.68} & \textbf{0.72 ± 0.01} & \textbf{0.97 ± 0.00} & \textbf{73.75 ± 0.41} & \textbf{98.85 ± 0.19} & \textbf{0.99 ± 0.00} & 0.99 ± 0.00 & \textbf{97.21 ± 0.44} & \textbf{99.87 ± 0.01} & \textbf{0.97 ± 0.00} & \textbf{0.96 ± 0.00} & \textbf{89.19 ± 0.22} \\
 & Ridge & 81.52 ± 0.73 & 0.71 ± 0.00 & 0.97 ± 0.00 & 72.36 ± 1.05 & 98.79 ± 0.17 & 0.99 ± 0.00 & 0.99 ± 0.00 & 97.09 ± 0.47 & 99.87 ± 0.01 & 0.97 ± 0.00 & 0.95 ± 0.00 & 89.07 ± 0.24 \\
\cmidrule{1-14}
\rowcolor{gray!15} \multirow{4}{*}{KPRPE $\rightarrow$ MagFace} & Baseline & 0.79 ± 0.18 & 0.10 ± 0.01 & 0.50 ± 0.01 & 1.02 ± 0.29 & 0.23 ± 0.09 & 0.09 ± 0.01 & 0.51 ± 0.10 & 1.13 ± 0.51 & 0.02 ± 0.01 & 0.08 ± 0.01 & 0.50 ± 0.01 & 1.06 ± 0.12 \\
 & Procrustes & 76.69 ± 1.43 & 0.65 ± 0.01 & \textbf{0.95 ± 0.00} & \textbf{58.00 ± 1.38} & 98.76 ± 0.24 & \textbf{0.99 ± 0.00} & 0.98 ± 0.00 & 96.60 ± 0.59 & 99.88 ± 0.02 & 0.97 ± 0.00 & 0.95 ± 0.00 & 89.00 ± 0.22 \\
 & Linear & \textbf{82.32 ± 0.70} & \textbf{0.66 ± 0.01} & 0.95 ± 0.00 & 51.74 ± 2.36 & \textbf{98.80 ± 0.21} & 0.99 ± 0.00 & \textbf{0.99 ± 0.00} & \textbf{96.77 ± 0.53} & \textbf{99.89 ± 0.02} & \textbf{0.97 ± 0.00} & \textbf{0.96 ± 0.00} & \textbf{89.17 ± 0.18} \\
 & Ridge & 73.37 ± 0.99 & 0.62 ± 0.01 & 0.93 ± 0.00 & 47.37 ± 1.75 & 98.75 ± 0.21 & 0.99 ± 0.00 & 0.99 ± 0.00 & 96.70 ± 0.54 & 99.88 ± 0.02 & 0.97 ± 0.00 & 0.95 ± 0.00 & 89.05 ± 0.22 \\
\specialrule{1.3pt}{2pt}{2pt}
\rowcolor{gray!15} \multirow{4}{*}{CLIP $\rightarrow$ ALIGN} & Baseline & 0.20 ± 0.13 & 0.08 ± 0.01 & 0.45 ± 0.01 & 0.37 ± 0.16 & 0.17 ± 0.07 & 0.08 ± 0.01 & 0.45 ± 0.06 & 0.73 ± 0.30 & 0.02 ± 0.01 & 0.09 ± 0.01 & 0.47 ± 0.00 & 0.82 ± 0.07 \\
 & Procrustes & 94.59 ± 0.89 & \textbf{0.66 ± 0.01} & \textbf{0.97 ± 0.00} & \textbf{57.29 ± 1.68} & 87.16 ± 0.64 & \textbf{0.76 ± 0.01} & \textbf{0.96 ± 0.01} & \textbf{53.64 ± 4.92} & 65.74 ± 0.29 & 0.62 ± 0.00 & \textbf{0.88 ± 0.00} & \textbf{27.46 ± 0.84} \\
 & Linear & 94.07 ± 0.60 & 0.62 ± 0.01 & 0.97 ± 0.00 & 50.32 ± 1.92 & \textbf{87.44 ± 0.21} & 0.75 ± 0.01 & 0.95 ± 0.01 & 43.78 ± 4.74 & \textbf{69.52 ± 0.29} & \textbf{0.66 ± 0.00} & 0.88 ± 0.00 & 24.70 ± 0.56 \\
 & Ridge & \textbf{94.65 ± 0.69} & 0.62 ± 0.01 & 0.97 ± 0.00 & 51.79 ± 2.23 & 87.39 ± 0.48 & 0.75 ± 0.01 & 0.95 ± 0.01 & 44.44 ± 4.91 & 69.31 ± 0.28 & 0.65 ± 0.00 & 0.88 ± 0.00 & 24.83 ± 0.66 \\
\cmidrule{1-14}
\rowcolor{gray!15} \multirow{4}{*}{BLIP-2 $\rightarrow$ LLaVA} & Baseline & 0.60 ± 0.22 & 0.09 ± 0.01 & 0.47 ± 0.00 & 0.87 ± 0.13 & 0.08 ± 0.08 & 0.10 ± 0.02 & 0.43 ± 0.06 & 0.16 ± 0.08 & 0.03 ± 0.01 & 0.08 ± 0.01 & 0.50 ± 0.00 & 1.04 ± 0.11 \\
 & Procrustes & 98.63 ± 0.36 & \textbf{0.90 ± 0.00} & \textbf{1.00 ± 0.00} & \textbf{91.59 ± 0.39} & 98.59 ± 0.46 & \textbf{0.94 ± 0.01} & \textbf{1.00 ± 0.00} & \textbf{93.42 ± 2.52} & 98.79 ± 0.08 & 0.76 ± 0.00 & \textbf{0.95 ± 0.00} & \textbf{61.12 ± 0.33} \\
 & Linear & 98.05 ± 0.56 & 0.85 ± 0.01 & 0.99 ± 0.00 & 82.54 ± 1.43 & \textbf{98.72 ± 0.45} & 0.93 ± 0.01 & 0.99 ± 0.00 & 91.58 ± 3.22 & \textbf{99.17 ± 0.07} & \textbf{0.77 ± 0.00} & 0.94 ± 0.00 & 57.35 ± 0.70 \\
 & Ridge & \textbf{98.68 ± 0.28} & 0.88 ± 0.01 & 0.99 ± 0.00 & 87.64 ± 1.31 & 98.67 ± 0.52 & 0.93 ± 0.01 & 0.99 ± 0.00 & 91.79 ± 3.65 & 99.13 ± 0.07 & 0.77 ± 0.00 & 0.94 ± 0.00 & 57.52 ± 0.67 \\
\cmidrule{1-14}
\rowcolor{gray!15} \multirow{4}{*}{DINOv2 $\rightarrow$ ViT} & Baseline & 0.75 ± 0.19 & 0.11 ± 0.00 & 0.46 ± 0.01 & 0.90 ± 0.09 & 0.70 ± 0.60 & 0.12 ± 0.04 & 0.50 ± 0.04 & 0.75 ± 0.27 & 0.04 ± 0.01 & 0.10 ± 0.01 & 0.50 ± 0.00 & 1.03 ± 0.04 \\
 & Procrustes & 81.55 ± 1.36 & 0.50 ± 0.01 & 0.90 ± 0.00 & 15.47 ± 1.49 & \textbf{85.36 ± 0.54} & \textbf{0.71 ± 0.01} & 0.87 ± 0.01 & 19.11 ± 3.75 & 77.42 ± 0.32 & 0.76 ± 0.00 & 0.82 ± 0.00 & 13.09 ± 0.32 \\
 & Linear & \textbf{84.76 ± 1.04} & \textbf{0.51 ± 0.00} & 0.91 ± 0.00 & \textbf{17.44 ± 1.43} & 85.24 ± 0.75 & 0.71 ± 0.01 & \textbf{0.89 ± 0.01} & \textbf{20.95 ± 3.64} & \textbf{81.46 ± 0.23} & \textbf{0.78 ± 0.00} & \textbf{0.83 ± 0.00} & \textbf{14.18 ± 0.19} \\
 & Ridge & 84.13 ± 0.94 & 0.51 ± 0.01 & \textbf{0.91 ± 0.00} & 17.23 ± 1.51 & 84.89 ± 0.76 & 0.70 ± 0.01 & 0.89 ± 0.01 & 20.92 ± 3.66 & 81.25 ± 0.26 & 0.78 ± 0.00 & 0.83 ± 0.00 & 14.14 ± 0.22 \\
\cmidrule{1-14}
\rowcolor{gray!15} \multirow{4}{*}{SAM $\rightarrow$ InternVL3} & Baseline & 0.71 ± 0.15 & 0.09 ± 0.00 & 0.52 ± 0.01 & 0.95 ± 0.15 & 0.68 ± 0.60 & 0.12 ± 0.03 & 0.53 ± 0.03 & 1.33 ± 0.38 & 0.04 ± 0.01 & 0.09 ± 0.00 & 0.54 ± 0.00 & 1.42 ± 0.11 \\
 & Procrustes & 11.52 ± 0.71 & 0.20 ± 0.00 & 0.71 ± 0.01 & 5.57 ± 0.71 & 6.66 ± 0.64 & 0.22 ± 0.01 & 0.67 ± 0.04 & 4.66 ± 1.66 & 4.36 ± 0.12 & 0.22 ± 0.00 & 0.65 ± 0.00 & 4.43 ± 0.20 \\
 & Linear & \textbf{63.64 ± 1.13} & \textbf{0.45 ± 0.01} & \textbf{0.83 ± 0.01} & \textbf{13.29 ± 1.80} & \textbf{46.18 ± 0.58} & \textbf{0.49 ± 0.01} & \textbf{0.81 ± 0.01} & \textbf{9.73 ± 3.01} & \textbf{39.76 ± 0.19} & \textbf{0.54 ± 0.00} & \textbf{0.72 ± 0.00} & \textbf{7.87 ± 0.23} \\
 & Ridge & 17.73 ± 0.65 & 0.23 ± 0.00 & 0.76 ± 0.01 & 8.06 ± 0.62 & 9.54 ± 0.59 & 0.24 ± 0.01 & 0.72 ± 0.03 & 4.52 ± 1.44 & 8.91 ± 0.23 & 0.29 ± 0.01 & 0.69 ± 0.00 & 5.47 ± 0.33 \\
\cmidrule{1-14}
\rowcolor{gray!15} \multirow{4}{*}{Florence-2 $\rightarrow$ Kosmos-2} & Baseline & 0.96 ± 0.17 & 0.11 ± 0.01 & 0.52 ± 0.01 & 1.23 ± 0.26 & 0.21 ± 0.05 & 0.10 ± 0.01 & 0.50 ± 0.04 & 0.72 ± 0.30 & 0.03 ± 0.01 & 0.08 ± 0.01 & 0.52 ± 0.00 & 1.06 ± 0.11 \\
 & Procrustes & 54.27 ± 0.71 & 0.41 ± 0.00 & 0.91 ± 0.00 & 21.88 ± 2.42 & 58.19 ± 1.45 & 0.57 ± 0.01 & 0.86 ± 0.02 & 11.86 ± 4.31 & 50.22 ± 0.38 & 0.58 ± 0.00 & 0.80 ± 0.00 & 12.02 ± 0.20 \\
 & Linear & \textbf{92.08 ± 0.66} & \textbf{0.62 ± 0.01} & \textbf{0.96 ± 0.00} & \textbf{45.16 ± 1.43} & \textbf{94.16 ± 0.71} & \textbf{0.79 ± 0.01} & \textbf{0.95 ± 0.01} & \textbf{40.49 ± 7.07} & \textbf{97.16 ± 0.14} & \textbf{0.81 ± 0.00} & \textbf{0.88 ± 0.00} & \textbf{29.35 ± 0.55} \\
 & Ridge & 84.07 ± 1.18 & 0.55 ± 0.01 & 0.95 ± 0.00 & 36.34 ± 1.65 & 90.02 ± 0.74 & 0.75 ± 0.01 & 0.93 ± 0.01 & 32.57 ± 5.87 & 95.27 ± 0.16 & 0.80 ± 0.00 & 0.88 ± 0.00 & 27.32 ± 0.48 \\
\bottomrule
\end{tabular}%
}
\end{table*}


\subsection{Face-Specific Models}
\subsubsection{Identification}

\begin{figure*}[t]
    \centering
    
    \includegraphics[width=0.49\textwidth]{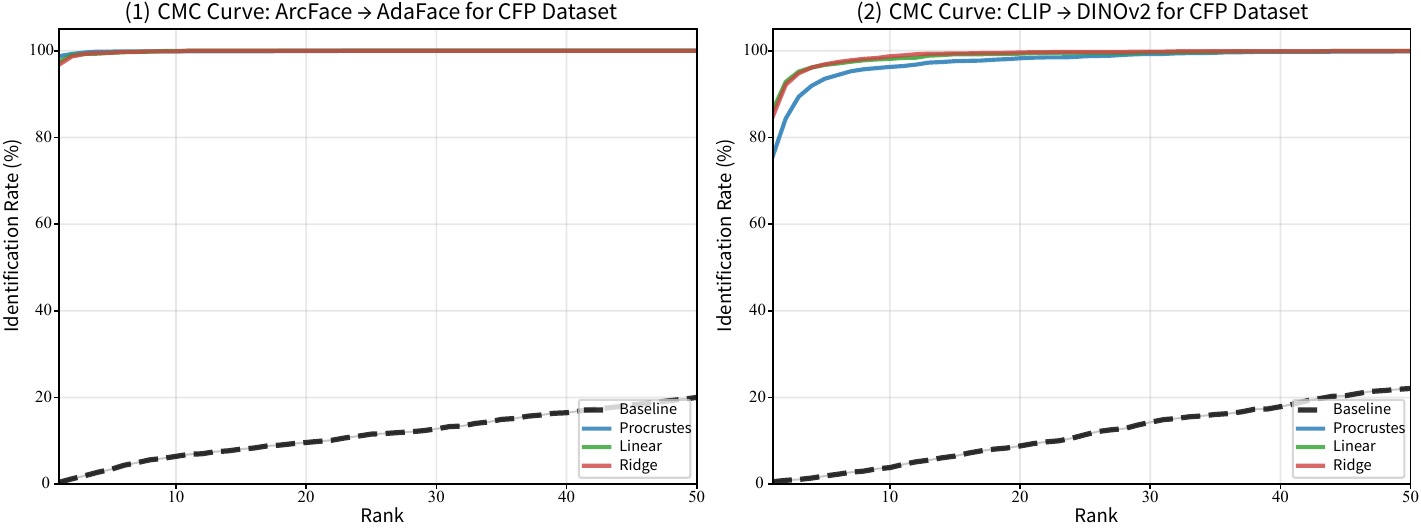}
    \hfill
    \includegraphics[width=0.49\textwidth]{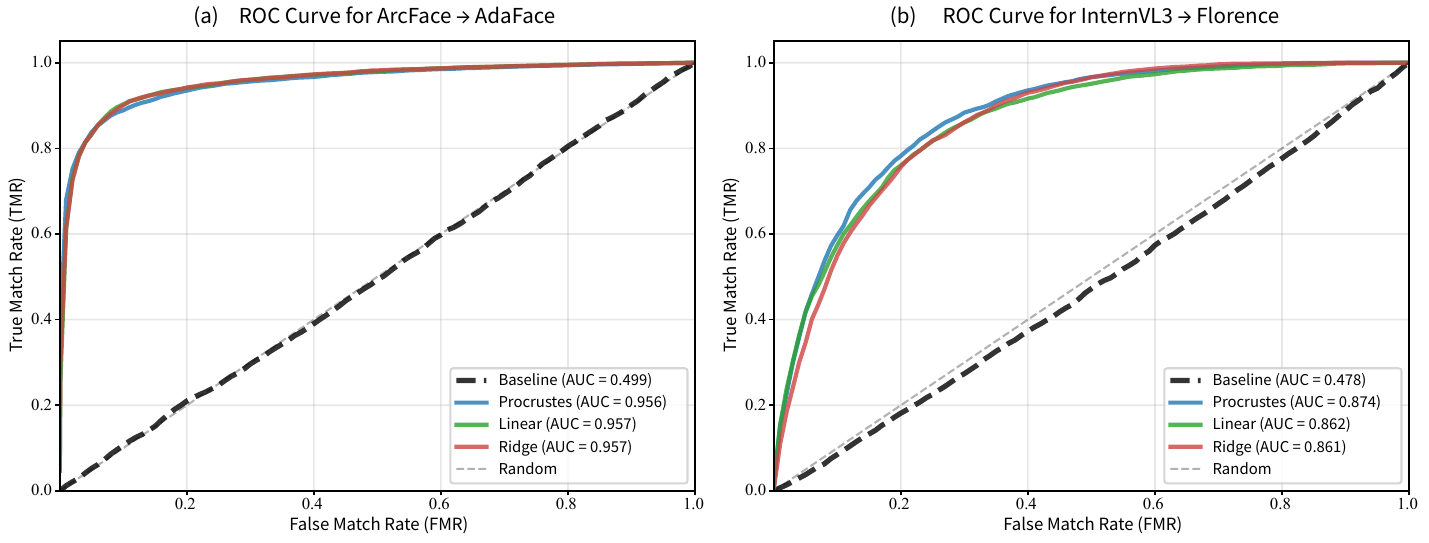}
    \caption{\textbf{Identification and verification curves.} \emph{Left}: CMC curves for face identification. \emph{Right}: ROC curves for face verification. Alignment substantially improves both over the unaligned baseline.}
    \label{fig:curves}
\end{figure*}

\begin{figure}[t]
  \centering
  \includegraphics[width=0.6\linewidth]{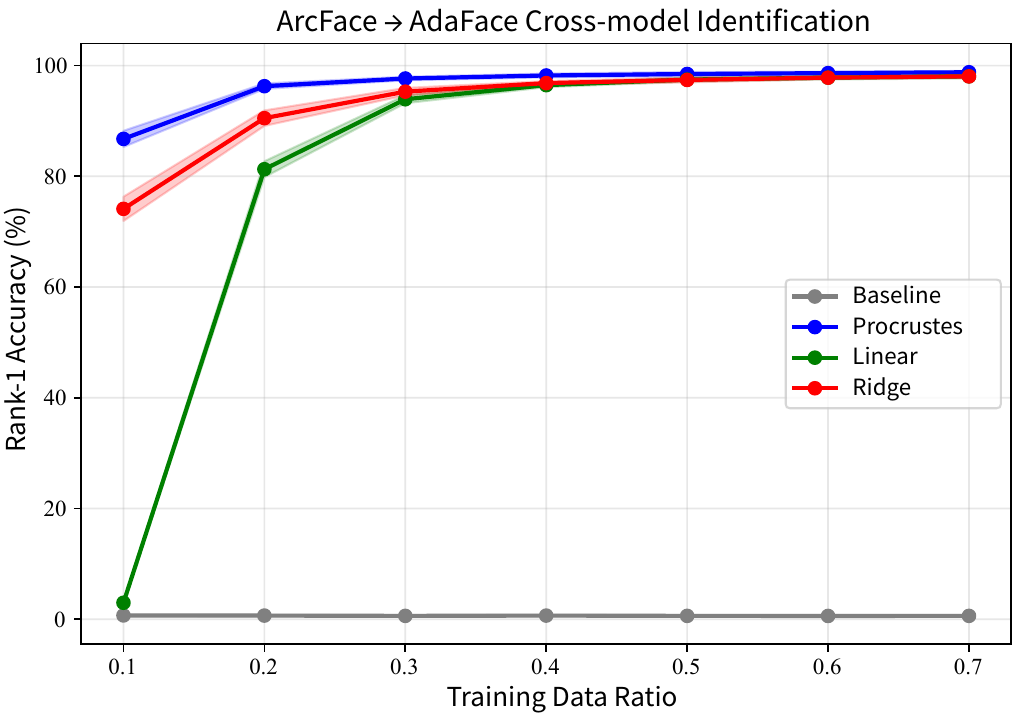}
  \caption{\textbf{Performance vs.\ training data.} Procrustes and Ridge remain robust with limited data, achieving over 70\% accuracy with only 10\% of the training set. Linear initially overfits but outperforms both methods once sufficient data is available.}
  \label{fig:trend}
\end{figure}

Table~\ref{tab:intra_main} presents intra-dataset identification results for face-specific model pairs. Without alignment, identification retrieval performance is poor, with mean Rank-1 below 1\% across all datasets. After alignment, performance improves substantially to average gains of +89.5\%, +98.85\%, and +99.88\% on CFP, LFW and WebFace datasets, respectively. Averaged across all model pairs and datasets, Linear achieves the best overall performance (given our choice of training data size, see Figure~\ref{fig:trend}), followed closely by Procrustes and Ridge. For all experiments, the regularization parameter $\alpha$ for Ridge is set to 0.1 based on preliminary tuning. As expected, closely related margin-based models such as ArcFace and AdaFace exhibit near-perfect alignment. Retrieval is slightly lower on KPRPE and MagFace indicating greater representational divergence. Figure~\ref{fig:curves} (left) shows CMC curves for ArcFace$\rightarrow$AdaFace on CFP where Linear Alignment results in 95\% of queries finding correct matches in the top-5 ranks versus 30\% at baseline.


\subsubsection{Verification}

Table~\ref{tab:intra_main} presents the corresponding intra-dataset verification results. Without alignment, performance is effectively random, with mean AUC $\approx\!\!0.50$ and TMR@1\%FMR around 1\% for all datasets. After alignment, verification improves substantially: mean AUC rises to 0.9585, 0.9868, and 0.9550, and TMR@1\%FMR increases to 64.20\%, 96.66\%, and 89.06\% on CFP, LFW and WebFace datasets, respectively. Averaged across all model pairs and datasets, Linear achieves the highest overall AUC, while Procrustes attains the best mean TMR@1\%FMR. Figure~\ref{fig:curves} (right) shows ROC curves for ArcFace $\rightarrow$ AdaFace on CFP. All alignment methods substantially outperform the unaligned baseline.

\subsection{Foundation Models}
\subsubsection{Identification}

Table~\ref{tab:intra_main} presents intra-dataset identification results for foundation model pairs. Without alignment, identification performance is near-zero, with baseline Rank-1 below 3\% across all datasets. After alignment, performance improves substantially with mean Rank-1 of 72.68\%, 72.02\%, and 68.70\% for CFP, LFW and WebFace, respectively. Linear still achieves the best overall performance, followed by Ridge and Procrustes. Performance varies systematically across model families: contrastive and multimodal alignment models exhibit stronger retrieval compatibility, whereas task-specific models such as SAM show the lowest alignment performance. In certain model pairs, Rank-1 goes up to 98\% as shown in the Table~\ref{tab:intra_main}. Figure~\ref{fig:curves} (left) illustrates CMC curves for CLIP$\rightarrow$DINOv2 on CFP.

\subsubsection{Verification}

Table~\ref{tab:intra_main} presents intra-dataset verification results for foundation model pairs. Without alignment, performance is near-random, with mean AUC $\approx\!\! 0.49$ and TMR@1\%FMR below 2\% across all datasets. After alignment, verification improves substantially: mean AUC rises to 0.8967, 0.8682, and 0.8208, and TMR@1\%FMR increases to 32.37\%, 31.62\%, and 21.75\% on CFP, LFW and WebFace datasets, respectively. Despite substantial improvement over baseline, foundation models achieve significantly lower verification performance than face-specific models revealing a critical gap between identification and verification tasks~\cite{decann2012can}. At restrictive security thresholds (FMR=0.1\%), TMR drops below 15\% for most foundation model pairs, indicating that alignment can improve ranking but cannot fundamentally tighten within-identity clustering required for high-security verification. Figure~\ref{fig:curves} (right) shows ROC curves for InternVL3$\rightarrow$Florence-2 on CFP.

\subsection{Cross-Dataset Generalization}
\begin{table}[t]
\centering
\scriptsize
\caption{\textbf{Cross-dataset generalization (CFP$\rightarrow$LFW).} Identification (R-1, mAP) and verification (AUC, TMR@1\%FMR) for a common set of pairs. Top block: face-specific; bottom block: foundation. \emph{Baseline} is the unaligned cross-model comparison.}
\label{tab:cross_combined}
\resizebox{\columnwidth}{!}{%
\begin{tabular}{ll cccc}
\toprule
\textbf{Model Pair} & \textbf{Method} & R-1 & mAP & AUC & TMR \\
\midrule
\rowcolor{gray!15} \multirow{4}{*}{ArcFace $\rightarrow$ AdaFace} & Baseline & 0.08 & 0.08 & 0.49 ± 0.00 & 0.89 ± 0.12 \\
 & Procrustes & \textbf{99.03} & \textbf{0.98} & \textbf{0.95 ± 0.00} & 88.83 ± 0.35 \\
 & Linear & 98.81 & 0.96 & 0.95 ± 0.00 & \textbf{88.85 ± 0.37} \\
 & Ridge & 98.86 & 0.97 & 0.95 ± 0.00 & 88.83 ± 0.37 \\
\cmidrule{1-6}
\rowcolor{gray!15} \multirow{4}{*}{AdaFace $\rightarrow$ KPRPE} & Baseline & 0.08 & 0.09 & 0.50 ± 0.01 & 0.99 ± 0.12 \\
 & Procrustes & \textbf{94.51} & \textbf{0.91} & 0.95 ± 0.00 & 87.49 ± 0.35 \\
 & Linear & 89.28 & 0.87 & \textbf{0.95 ± 0.00} & \textbf{88.36 ± 0.34} \\
 & Ridge & 88.76 & 0.86 & 0.95 ± 0.00 & 88.02 ± 0.34 \\
\cmidrule{1-6}
\rowcolor{gray!15} \multirow{4}{*}{KPRPE $\rightarrow$ MagFace} & Baseline & 0.04 & 0.09 & 0.50 ± 0.00 & 1.03 ± 0.06 \\
 & Procrustes & \textbf{95.12} & \textbf{0.92} & 0.95 ± 0.00 & 88.25 ± 0.39 \\
 & Linear & 86.78 & 0.84 & \textbf{0.95 ± 0.00} & \textbf{88.40 ± 0.38} \\
 & Ridge & 88.39 & 0.85 & 0.95 ± 0.00 & 88.28 ± 0.37 \\
\specialrule{1.3pt}{2pt}{2pt}
\rowcolor{gray!15} \multirow{4}{*}{CLIP $\rightarrow$ ALIGN} & Baseline & 0.04 & 0.08 & 0.36 ± 0.00 & 0.23 ± 0.03 \\
 & Procrustes & \textbf{64.66} & \textbf{0.62} & 0.95 ± 0.00 & 51.14 ± 1.33 \\
 & Linear & 60.38 & 0.59 & 0.96 ± 0.00 & \textbf{55.68 ± 1.11} \\
 & Ridge & 61.69 & 0.60 & \textbf{0.96 ± 0.00} & 55.43 ± 0.71 \\
\cmidrule{1-6}
\rowcolor{gray!15} \multirow{4}{*}{BLIP-2 $\rightarrow$ LLaVA} & Baseline & 0.01 & 0.09 & 0.44 ± 0.01 & 0.09 ± 0.03 \\
 & Procrustes & \textbf{97.09} & \textbf{0.91} & \textbf{1.00 ± 0.00} & \textbf{95.12 ± 0.28} \\
 & Linear & 94.51 & 0.87 & 0.99 ± 0.00 & 90.74 ± 0.70 \\
 & Ridge & 96.10 & 0.89 & 1.00 ± 0.00 & 93.82 ± 0.43 \\
\cmidrule{1-6}
\rowcolor{gray!15} \multirow{4}{*}{DINOv2 $\rightarrow$ ViT} & Baseline & 0.08 & 0.09 & 0.42 ± 0.00 & 0.27 ± 0.06 \\
 & Procrustes & \textbf{56.10} & \textbf{0.58} & \textbf{0.91 ± 0.00} & 21.39 ± 1.03 \\
 & Linear & 50.95 & 0.55 & 0.90 ± 0.00 & 21.68 ± 1.31 \\
 & Ridge & 51.70 & 0.55 & 0.91 ± 0.00 & \textbf{21.78 ± 1.58} \\
\cmidrule{1-6}
\rowcolor{gray!15} \multirow{4}{*}{SAM $\rightarrow$ InternVL3} & Baseline & 0.03 & 0.08 & 0.31 ± 0.00 & 0.28 ± 0.08 \\
 & Procrustes & 0.42 & 0.11 & 0.65 ± 0.00 & 1.67 ± 0.12 \\
 & Linear & \textbf{4.07} & \textbf{0.20} & \textbf{0.81 ± 0.00} & \textbf{8.45 ± 0.41} \\
 & Ridge & 0.88 & 0.15 & 0.71 ± 0.00 & 3.16 ± 0.24 \\
\cmidrule{1-6}
\rowcolor{gray!15} \multirow{4}{*}{Florence-2 $\rightarrow$ Kosmos-2} & Baseline & 0.06 & 0.10 & 0.45 ± 0.00 & 0.25 ± 0.05 \\
 & Procrustes & 14.84 & 0.33 & 0.89 ± 0.00 & 24.90 ± 1.39 \\
 & Linear & \textbf{58.59} & \textbf{0.63} & \textbf{0.96 ± 0.00} & \textbf{60.97 ± 0.77} \\
 & Ridge & 48.52 & 0.57 & 0.96 ± 0.00 & 47.93 ± 1.15 \\
\bottomrule
\end{tabular}%
}
\end{table}

\subsubsection{Identification}
Table~\ref{tab:cross_combined} reports cross-dataset results (CFP$\rightarrow$LFW) for a common set of model pairs, evaluated on both identification and verification. Considering identification, foundation-model pairs drop from their intra-dataset level (72.68\% Rank-1 on CFP) to a mean Rank-1 of 39.21\% (mAP 0.454). The relative performance ordering of alignment methods mirrors the intra-dataset scenario. This degradation reflects distribution shift: alignments learned under CFP’s controlled conditions do not fully transfer to LFW’s unconstrained, in-the-wild images. Nevertheless, performance remains markedly above the unaligned baseline, indicating that the geometric relationships between models exhibit partial domain invariance and that the core alignment structure generalizes beyond the training distribution. Face-specific pairs, by contrast, generalize strongly (mean Rank-1 95.21\%, mAP 0.923).

\subsubsection{Verification}

The same Table~\ref{tab:cross_combined} reports cross-dataset verification (CFP$\rightarrow$LFW) for the same pairs.
Without alignment, face-specific model performance is near-random. After alignment, verification transfers robustly across datasets: face-specific pairs reach mean AUC 0.950 (TMR@1\%FMR 88.21\%) and foundation pairs mean AUC 0.858 (TMR@1\%FMR 37.63\%). Linear achieves the best overall performance, with Procrustes and Ridge performing comparably. This indicates that the transformation required to align embeddings is nearly domain-invariant and can retain identity-discriminative features that remain consistent across dataset-specific variations in capture conditions, image quality, and demographics.

\section{Discussion}
\subsection{Why Linear Transformations Suffice}
A central question raised by our results is why simple linear transformations achieve substantial cross-model alignment, despite differences in training data, architectures, and training objectives. We propose a geometric interpretation grounded in manifold learning theory.

Our findings suggest that, when applied to face recognition, different models learn embeddings that lie on approximately the same low-dimensional identity manifold in semantic space. This manifold captures the intrinsic degrees of freedom governing facial appearance (e.g., pose, expression, illumination)~\cite{o2018face}. While training procedures of DNN models may yield different parameterizations of this structure, the underlying geometry appears largely shared. Although this identity manifold may be globally curved, it is locally well-approximated by its tangent space. Because standard face recognition benchmarks sample a bounded region of this manifold, the data effectively occupy such a local neighborhood. In this regime, cross-model alignment reduces to estimating a linear map between two tangent-space representations of the same intrinsic structure. Thus, simple linear transformations suffice: they approximate a coordinate change between local linearizations of a shared geometric manifold. In the supplementary material, we further support this empirically: the linear map improves embedding-level similarity (RSA and CKA), while fitting non-linear quadratic and MLP maps yields little additional gain.

\begin{figure}[t]
    \centering
  \includegraphics[width=\linewidth]{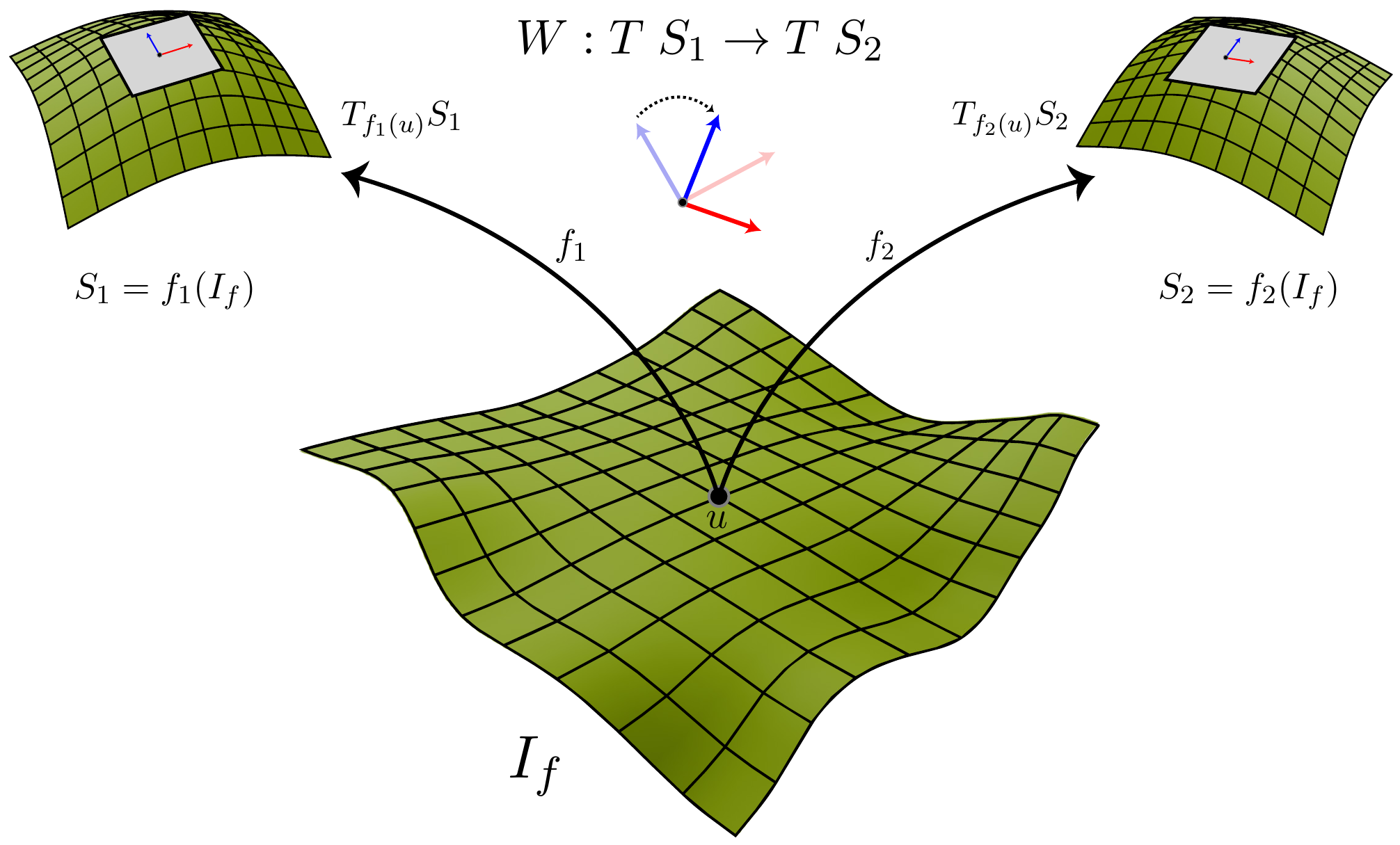}
  \caption{\textbf{Geometric interpretation of cross-model alignment.} A face image $u \in \mathcal{I}_f$ lies on a low-dimensional identity manifold. Two embeddings $f_1$ and $f_2$ map $\mathcal{I}_f$ into feature spaces $\mathcal{F}_1$ and $\mathcal{F}_2$, producing submanifolds $S_1$ and $S_2$. Locally, these are approximated by tangent spaces, and the learned linear map $W$ aligns $T_{f_1(u)}S_1$ to $T_{f_2(u)}S_2$ as a coordinate transformation.}
  \label{fig:manifold}
  \vspace{-6pt}
\end{figure}

More precisely, we may postulate our result as an alignment of the embeddings of a $d_f$-dimensional submanifold of face images $\mathcal{I}_f$ within the set of generic images $\mathcal{I}$, under two embeddings $f_1$ and $f_2$ as demonstrated in Figure~\ref{fig:manifold}. An ideal nonlinear alignment would have $\smash{\varphi(f_1(u))\!=\!f_2(u)}$ for any $\smash{u\!\in\!\mathcal{I}_f},$ i.e., $\smash{f_2^{\scriptscriptstyle-1}\circ\varphi\circ f_1}$ is the identity map. Assuming smoothness of the alignment mapping $\varphi$, we may linearize it in the neighborhood of $\bar{u}$ which is a mean face image, with the embedding $f_1(\bar{u})$ and $f_2(\bar{u})$ estimated by the respective training set means ($\mu_X$ and $\mu_Y$). Our linear map $W$ can be interpreted by the differential $D\varphi$ (Jacobian matrix) that aligns the image of tangent space $T_u \mathcal{I}_f$ under the embeddings as affine subspaces $T_{\scriptstyle f_1(u)} S_1$ and $T_{\scriptstyle f_2(u)} S_2$ of the corresponding feature spaces for the embedding $\smash{S_i=f_i(\mathcal{I}_f})$. The linear map directly finds the best $D\varphi$ assuming the samples span the entire feature spaces, whereas the ridge regression offers an alignment when the samples may span only a subspace of the feature spaces, as these embeddings are typically not diffeomorphisms. Under the assumption that the embedding Hilbert spaces learned a good approximation of the same kernel function $\smash{K(u_1,u_2)\approx \left<f_1(u_1),f_1(u_2)\right>\approx\left<f_2(u_1),f_2(u_2)\right>},$ $W$ can be interpreted as the orthogonal transformation $W\in O(D)$ that aligns the images of the tangent space of $\mathcal{I}_f$ under the two embeddings (note that results are not influenced by how the vectors outside of the tangent spaces are rotated or reflected).

\begin{figure*}[t]
    \centering
    
    \begin{subfigure}{0.37\textwidth}
        \centering
        \includegraphics[width=\linewidth]{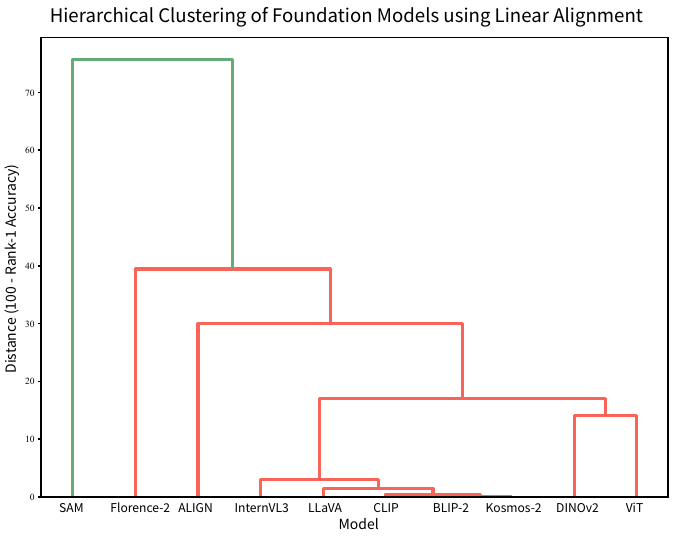}
        \caption{Model hierarchy derived from symmetrized cross-model alignment.}
        \label{fig:dendogram}
    \end{subfigure}
    \hfill
    \begin{subfigure}{0.5\textwidth}
        \centering
        \includegraphics[width=\linewidth]{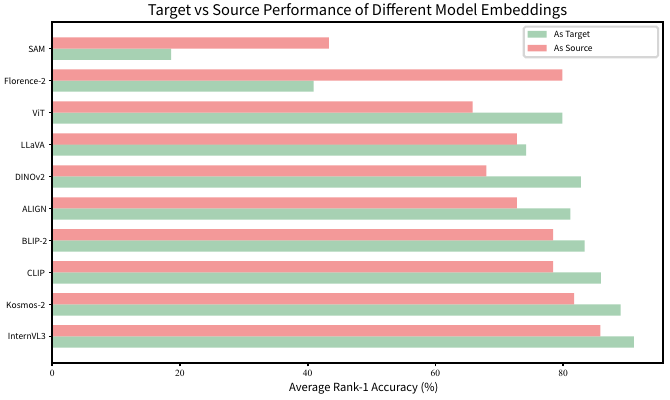}
        \caption{Average incoming and outgoing alignment across models. Differences reflect directional asymmetry.}
        \label{fig:asymmetry}
    \end{subfigure}
    
    \caption{\textbf{Analysis of cross-model compatibility.}}
    \label{fig:structure}
    \vspace{-6pt}
\end{figure*}

\subsection{Model Organization and Hierarchy}
To understand global structure in cross-model compatibility, we performed hierarchical clustering on the symmetrized alignment matrix (average of $\smash{A \!\rightarrow\! B}$ and $\smash{B\! \rightarrow\! A}$). The resulting dendrogram shown in Figure~\ref{fig:dendogram} reveals a consistent organization of models into coherent families, despite no architectural information being used in clustering. SAM separates first, reflecting its segmentation-focused optimization. Florence-2, a unified multi-task sequence-to-sequence vision model, branches next. ALIGN forms the third major branch. The remaining models cluster tightly, suggesting shared representational structure. Contrastive and multimodal alignment models (InternVL3, LLaVA, BLIP-2, CLIP) form one group, consistent with their shared objective of visual–text alignment. DINOv2 (self-supervised) and supervised ViT appear as a related subcluster. More importantly, this hierarchy remains stable across datasets, indicating representational convergence rather than dataset-specific effects. This organization has several practical implications. Beyond descriptive analysis, the hierarchy provides a structural map of embedding compatibility: it enables model substitution or simulation within clusters, offers a principled way to identify the closest open-source alternatives to proprietary systems, facilitates interoperability and ensemble design, and allows newly introduced models to be positioned within an existing geometric landscape without requiring architectural inspection or retraining. For biometric pipelines, this also offers concrete guidance on model choice: contrastive and vision-language foundation models align most readily with face-recognition embeddings, whereas segmentation-pretrained models such as SAM are the least compatible identity targets.


\subsection{Directional Asymmetry and Target-Space Compatibility}

Although many model pairs exhibit high mutual alignment, we observe substantial directional asymmetry (mean Rank-1 deviation of 15.7\%, with extremes exceeding 49\%), indicating that mapping from model $A$ to $B$ can differ markedly from the reverse direction. Part of this asymmetry can be attributed to embedding dimensionality gaps, suggesting potential information compression when aligning into lower-dimensional targets (e.g., SAM at 256D). However, dimensionality alone does not fully explain the phenomenon, as Florence-2 (768D) also appears as an outlier. This indicates that retrieval performance depends not only on dimensionality but also on the geometric quality of the target embedding, specifically how well identities are separated within that space. This also explains why face-specific models yield stronger retrieval performance: their training objectives explicitly enforce identity-discriminative structure. In contrast, models trained for segmentation may not prioritize identity separation, making them harder alignment targets. To further examine this effect and demonstrate the asymmetry, we compute both average incoming alignment (how well other models align to a given embedding space) and outgoing alignment (how well a model aligns to others). On the CFP dataset (Figure~\ref{fig:asymmetry}), InternVL3, Kosmos-2, and CLIP exhibit high incoming and outgoing compatibility. In contrast, SAM and Florence-2 show consistently lower incoming alignment, suggesting that their embedding geometries are more specialized or less identity-discriminative, even if they can align reasonably well to other models as sources.

\section{Conclusion}
In this work, we investigated whether different DNN models encode facial identity in fundamentally incompatible ways or use different parameterizations of a shared structure. Through systematic evaluation of linear alignment methods across domain-specific and domain-agnostic DNNs on identification and verification tasks within and across datasets, we find evidence for the latter. Without alignment, cross-model retrieval accuracy is close to random, averaging 2\%. Simple linear transformations achieve substantial cross-model retrieval accuracy (97\% for face-specific models, 71\% for foundation models), with learned transformations generalizing across datasets. Hierarchical analysis reveals stable model families, and directional asymmetry shows that target space quality affects retrieval performance.

These findings have dual implications. Positively, cross-model compatibility enables heterogeneous ensembles and interoperability: a template enrolled in one system can be reused in another through a learned map operating purely on embeddings, without re-acquiring the subject's image. Negatively, our results challenge biometric template protection schemes assuming system-specific embeddings provide inherent revocability. Future work should investigate whether this alignment phenomenon extends to other biometric modalities (iris, fingerprint) and fine-grained visual categories more broadly. Additionally, exploring cross-model alignment in the context of specific adversarial attacks would shed light on vulnerabilities introduced by geometric compatibility.

\bibliographystyle{ieee}
\bibliography{main}
\end{document}